\documentclass[preprint,12pt]{elsarticle}

\usepackage{amssymb}
\usepackage{amsmath}
\usepackage{algorithmic}
\usepackage[linesnumbered,ruled,vlined]{algorithm2e}
\usepackage{booktabs} 
\usepackage{hyperref}
\hypersetup{colorlinks=true, citecolor=blue, linkcolor=blue, urlcolor=blue}
\usepackage{multirow}
\usepackage{xcolor} 

\journal{Engineering Structures}

\begin{document}

\begin{frontmatter}



\title{Staircase Cascaded Fusion of Lightweight Local Pattern Recognition and Long-Range Dependencies for Structural Crack Segmentation} 



\author[tut_1,tut_2]{Hui Liu}
\ead{liuhui1109@stud.tjut.edu.cn}
\affiliation[tut_1]{organization={Engineering Research Center of Learning-Based Intelligent System (Ministry of Education)},
          city={Tianjin},
          postcode={300384}, 
           country={China}}
\affiliation[tut_2]{organization={Key Laboratory of Computer Vision and System (Ministry of Education)},
          city={Tianjin},
          postcode={300384}, 
           country={China}}
\author[tut_1,tut_2]{Chen Jia\corref{cor1}}
\ead{jiachen@email.tjut.edu.cn}
\cortext[cor1]{Corresponding author}

\author[tut_1,tut_2]{Fan Shi}
\ead{shifan@email.tjut.edu.cn}

\author[tut_1,tut_2]{Xu Cheng}
\ead{xu.cheng@ieee.org}

\author[tut_1,tut_2]{Mianzhao Wang}
\ead{wmz@email.tjut.edu.cn}

\author[tut_1,tut_2]{Shengyong Chen}
\ead{sy@ieee.org}

\author[tcu_1,tcu_2]{Yang Lv}
\ead{lvyangtju@163.com}
\affiliation[tcu_1]{organization={Tianjin Key Laboratory of Civil Structure Protection and Reinforcement},
          city={Tianjin},
          postcode={300384}, 
           country={China}}
\affiliation[tcu_2]{organization={Tianjin Chengjian University},
          city={Tianjin},
          postcode={300384}, 
           country={China}}

\begin{abstract}
Accurately segmenting structural cracks at the pixel level remains a major hurdle, as existing methods fail to integrate local textures with pixel dependencies, often leading to fragmented and incomplete predictions. Moreover, their high parameter counts and substantial computational demands hinder practical deployment on resource-constrained edge devices. To address these challenges, we propose CrackSCF, a Lightweight Cascaded Fusion Crack Segmentation Network designed to achieve robust crack segmentation with exceptional computational efficiency. We design a lightweight convolutional block (LRDS) to replace all standard convolutions. This approach efficiently captures local patterns while operating with a minimal computational footprint. For a holistic perception of crack structures, a lightweight Long-range Dependency Extractor (LDE) captures global dependencies. These are then intelligently unified with local patterns by our Staircase Cascaded Fusion Module (SCFM), ensuring the final segmentation maps are both seamless in continuity and rich in fine-grained detail. To comprehensively evaluate our method, this paper created the challenging TUT benchmark dataset and evaluated it alongside five other public datasets. The experimental results show that the CrackSCF method consistently outperforms the existing methods, and it demonstrates greater robustness in dealing with complex background noise. On the TUT dataset, CrackSCF achieved 0.8382 on F1 score and 0.8473 on mIoU, and it only required 4.79M parameters.
\end{abstract}

\begin{graphicalabstract}
    \includegraphics[width=\textwidth]{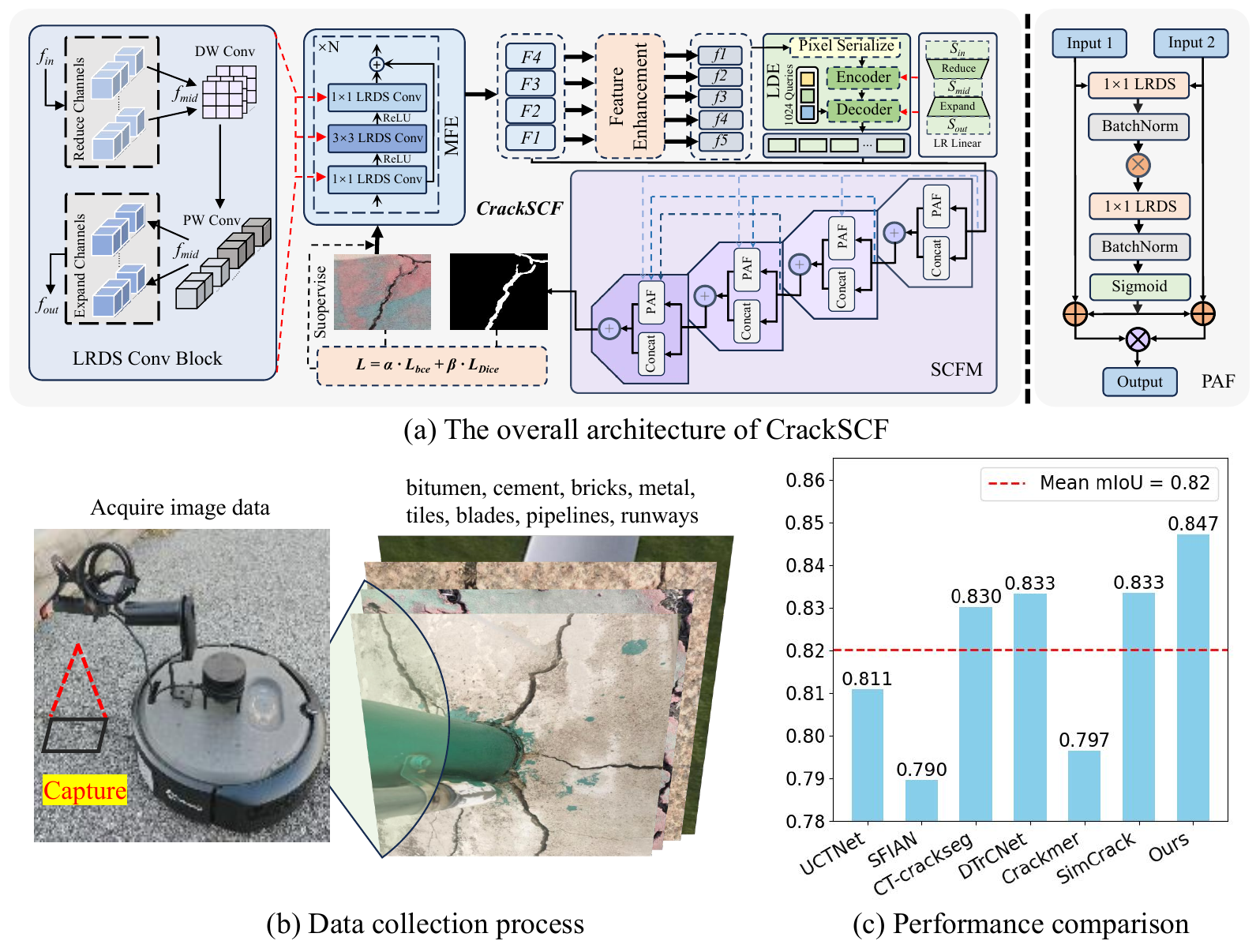}
\end{graphicalabstract}

\begin{highlights}
\item Research highlight 1

This paper proposed a novel CrackSCF network for structural crack segmentation. By introducing a lightweight convolution block, a lightweight long-range dependency extractor, and a staircase cascaded fusion module, this network effectively distinguishes between background and crack pixels with low computational resources, resulting in high-quality segmentation maps.

\item Research highlight 2

This paper collected the TUT dataset, which includes structural cracks in eight different scenarios, providing a more comprehensive evaluation of the network's performance.

\item Research highlight 3

Comprehensive experiments conducted on five public datasets and the TUT benchmark dataset demonstrate that the proposed CrackSCF network outperforms existing SOTA methods.

\end{highlights}


\end{frontmatter}

\section{Introduction}
\label{sec:introuction}

Cracks are common defects in structures such as pavements and buildings, which can lead to serious safety issues. Thus, automated detection on edge control devices is crucial for ensuring production safety \cite{ali2022structural, chen2023classification, reta2025structural, liu2025neighborhood, mihaylov2025rapid}. Early automated detection methods used traditional image processing techniques for crack extraction \cite{zhou2006wavelet, yamaguchi2008image, oliveira2009automatic, teng2024automated}. While these methods are easy to implement, their performance in complex scenarios still needs improvement.


In recent years, benefiting from the strong local inductive capability of Convolutional Neural Networks (CNN), CNN-based networks such as MST-Net \cite{qu2021crack} and FcaHRNet \cite{chu2024cascade} have shown efficient feature extraction capabilities in structural crack segmentation tasks. The spatial invariance of CNNs allows them to detect cracks anywhere in an image by learning similar features at different locations. In labeled images, crack labels are long, narrow structures of white pixels, while the background is large black areas. This challenges the model to distinguish between thin cracks and the background. Nevertheless, networks based on CNNs face challenges in capturing long-range dependencies between crack pixels, leading to segmentation discontinuities.

The success of Vision Transformer (ViT) \cite{dosovitskiy2020image} and Swin Transformer \cite{liu2021swin} has demonstrated the advantages of Transformer \cite{vaswani2017attention}, with their strong sequence processing capabilities, in handling visual tasks \cite{yun2023spectr, zhu2020deformable}. Unlike CNNs, transformers can capture dependencies between any pixels when processing images \cite{geng2023dual, tan2025cds}, making them more advantageous in handling complex topologies. To harness the advantages of both CNNs and transformers effectively, researchers have proposed networks such as CATransUNet \cite{chu2024transformer}, MFAFNet \cite{dong2024mfafnet}, and Crackmer \cite{xiang2023crack}, which integrate local  and long-range information through U-shaped connections or residual connections.

\begin{figure}
  \centering
  \includegraphics[width=0.6\textwidth]{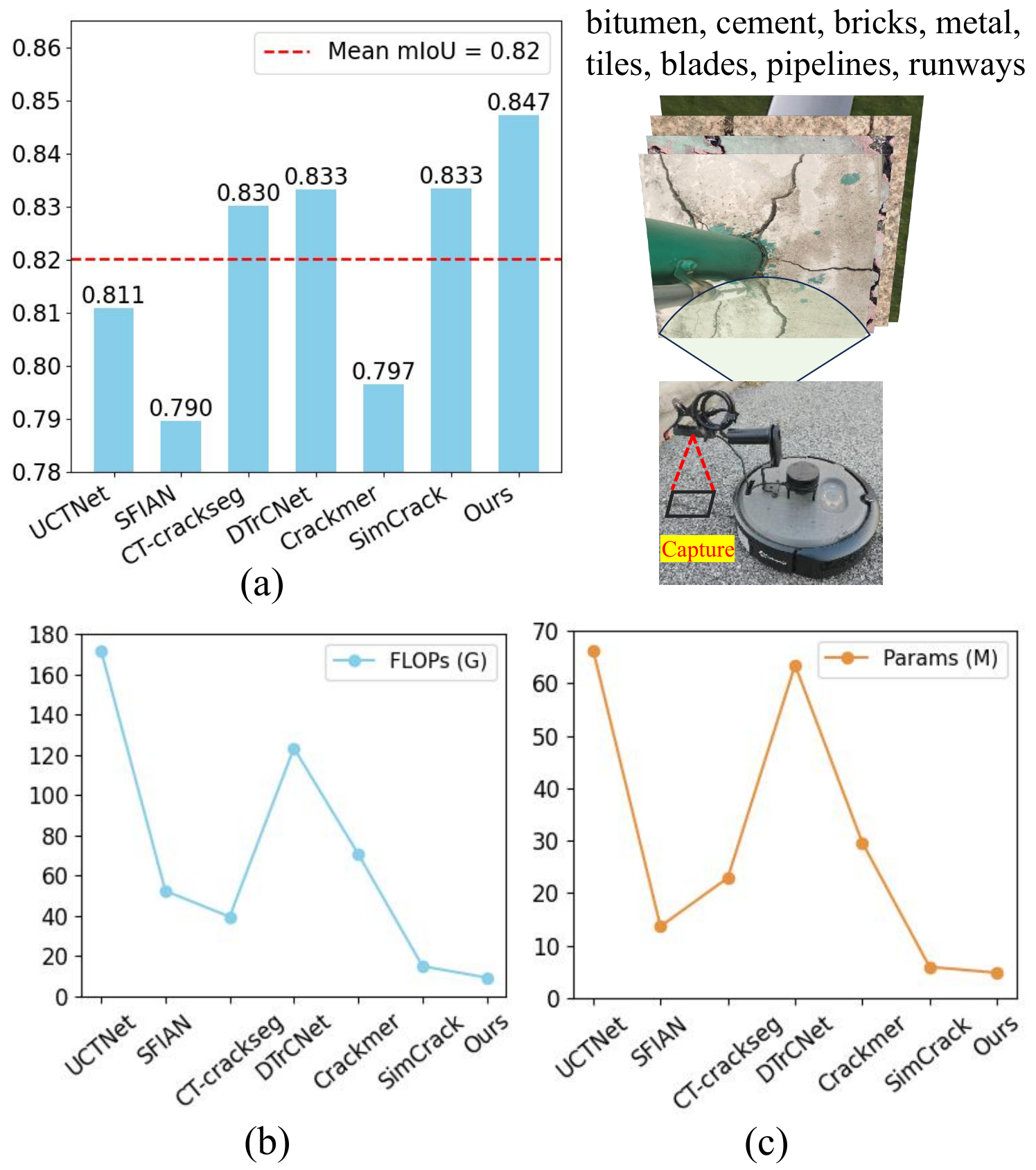}
  \caption{Performance comparison of our proposed method with UCTNet \cite{wang2022uctransnet}, SFIAN \cite{cheng2023selective}, CT-crackseg \cite{tao2023convolutional}, DTrCNet \cite{xiang2023crack}, Crackmer \cite{wang2024dual} and Simcrack \cite{jaziri2024designing} on the TUT dataset in terms of mIoUs, FLOPs and Params metrics.}
  \label{fig:Intro TUT result}
  \vspace{-0.6cm}
\end{figure}

Despite the aforementioned methods combining CNN and Transformer have improved performance in pixel-level crack segmentation tasks, they still have some limitations. The different information fusion strategies significantly affect the segmentation performance of CNN-Transformer networks. For instance, UCTNet \cite{wang2022uctransnet} and CT-crackseg \cite{tao2023convolutional} use channel concatenation to fuse local patterns and long-range dependences information. However, concatenation merely stacks the feature information together without deeper feature interaction, failing to understand their correlation and complementarity, resulting in poor segmentation performance on fine cracks and susceptibility to background noise. Additionally, DTrCNet \cite{xiang2023crack} and Crackmer \cite{wang2024dual} use channel attention to fuse information, which can enhance focus on certain features but neglects spatial information and pixel details in the feature maps, impacting the model's overall performance. As shown in Figure \ref{fig:Intro TUT result}(a), due to these shortcomings, these four methods fail to achieve the best performance.

Most networks, in pursuit of performance, overlook the significant parameter and computational costs brought by numerous convolution operations, this makes them challenging to deploy on resource-constrained control devices. As shown in Figure \ref{fig:Intro TUT result}(b) and Figure \ref{fig:Intro TUT result}(c), networks like SFIAN \cite{cheng2023selective}, Crackmer \cite{wang2024dual}, and CT-crackseg\cite{tao2023convolutional} have very high computational FLOPs and parameter counts, indicating that they are difficult to implement on resource-limited devices like smartphones and drones, significantly diminishing their practicality.

To address these issues, this paper proposes a staircase cascaded fusion network that generates high-quality segmentation maps with low computational resources while effectively suppressing background noise. Specifically, to reduce the number of parameters and computational burden, a lightweight convolution block called LRDS is introduced. This module utilizes the principle of low-rank approximation to effectively reduce the computational resources required by convolutional layers and captures spatial structures efficiently in the low-rank space through depthwise convolution, maintaining strong feature perception. The proposed architecture also includes a lightweight Long-range Dependency Extractor (LDE), which uses a lightweight deformable attention mechanism to effectively capture irregular long-range dependencies between crack pixels with minimal computational demands. Additionally, to better leverage the complementarity and correlation of feature maps while ensuring that critical detail information is not missed, the Staircase Cascaded Fusion Module (SCFM) is incorporated. This module generates feature maps layer by layer and effectively fuses them using a pixel attention mechanism that finely captures local and long-range information along with a channel concatenation mechanism, comprehensively combining local texture information and pixel dependencies.

To comprehensively evaluate the structural crack segmentation performance of the model in various complex environments, this paper also collected a dataset named TUT, which contains 1408 RGB images. Existing public datasets like DeepCrack \cite{liu2019deepcrack} and Crack500 \cite{yang2019feature} include only 1-2 types of scenarios and have relatively simple crack shapes, which are insufficient for thoroughly testing the network's performance. The TUT dataset includes a broader range of real structural crack image scenarios, covering plastic runways, bricks, tiles, cement, bitumen, generator blades, metal materials, and underground pipelines. With such a diverse set of image scenarios, this dataset effectively evaluates the network's generalization and robustness.  Furthermore, the images feature occlusions, highly complex backgrounds, and uneven lighting, which further enhances the datasets diversity.

\section{Methodology}
\label{section:Methodology}

\subsection{Network Architecture}

\begin{figure*}[!t]
  \centering
  \includegraphics[width=\textwidth]{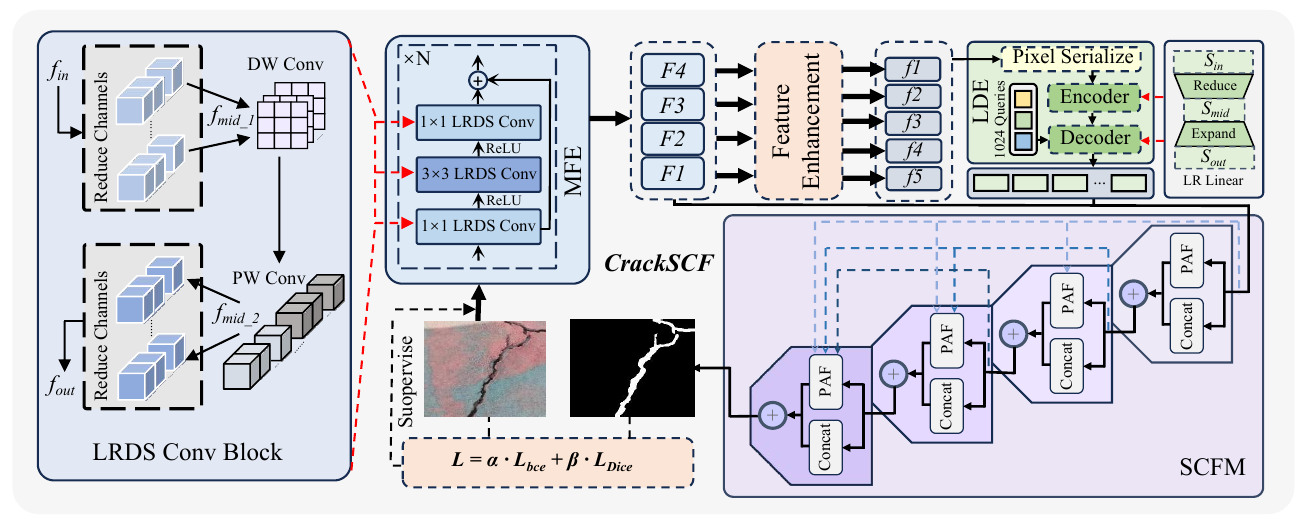}
  \caption{General Architecture Diagram of CrackSCF Network. Crack images are first input into the MFE, generating four feature maps. After enhancement, five feature maps with unified channel numbers are obtained. The LDE processes these maps to acquire pixel sequences rich in long-range correlations. The SCFM then processes these sequences and the four layers of feature maps from the MFE. Through four stages, resulting in the segmentation output.}
  \label{fig:Network_Arch}
\end{figure*}

As shown in Figure \ref{fig:Network_Arch}, the CrackSCF network has three main components: a lightweight Multi-scale Feature Extractor (MFE), a lightweight Long-range Dependency Extractor (LDE), and a Staircase Cascaded Fusion Module (SCFM). The overall architecture of the network is shaped like a staircase, with feature maps of different scales progressively fused and upsampled. In this process, the advantages of CNNs and Transformers in feature extraction complement each other.

Specifically, for a given dataset, the crack images are first input into the MFE module, which is divided into four stages. In each stage, the extracted feature maps undergo a Feature Enhancement stage. Then, the LDE module processes these enhanced feature maps to obtain pixel sequences rich in long-range dependencies. Subsequently, the SCFM module receives these sequences along with the four layers of feature maps extracted by the MFE. After processing through four stages, the resolution is doubled at each stage while the number of channels is halved, ultimately resulting in the segmentation result.

\begin{figure}[h]
  \centering
  \includegraphics[width=0.7\textwidth]{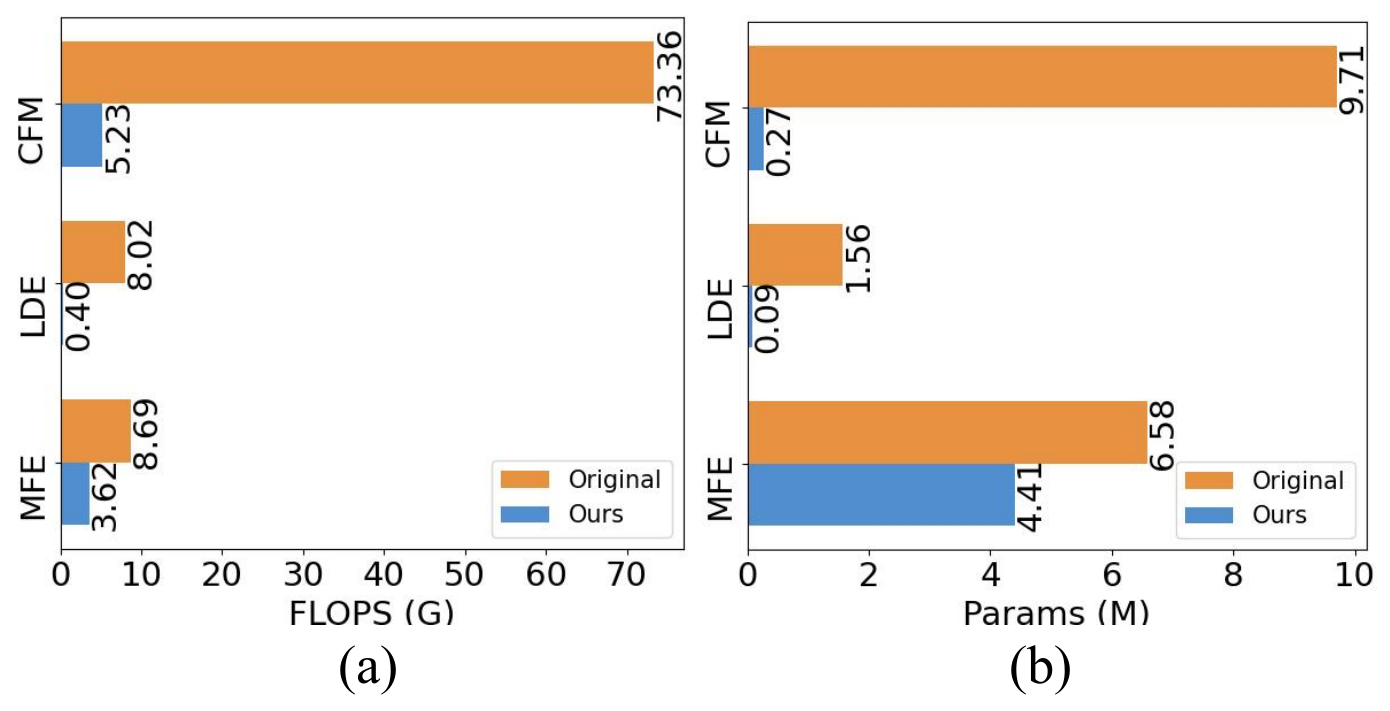}
  \caption{Comparison of FLOPs and Params for each module using the original convolution and our proposed LRDS convolution block.}
  \label{fig:FLOPS_Params_Ana}
  \vspace{-0.5cm}
\end{figure}

Importantly, as shown in Figure \ref{fig:FLOPS_Params_Ana}, using the original convolution operations results in high FLOPs and Params, especially in the SCFM module, reaching 73.36G and 9.71M, respectively. Replacing all convolutions with LRDS blocks significantly reduced FLOPs and Params, demonstrating the effectiveness of LRDS in model lightweighting.

\subsection{Lightweight Multi-scale Feature Extractor}

\begin{figure}[!t]
  \centering
  \includegraphics[width=0.6\textwidth]{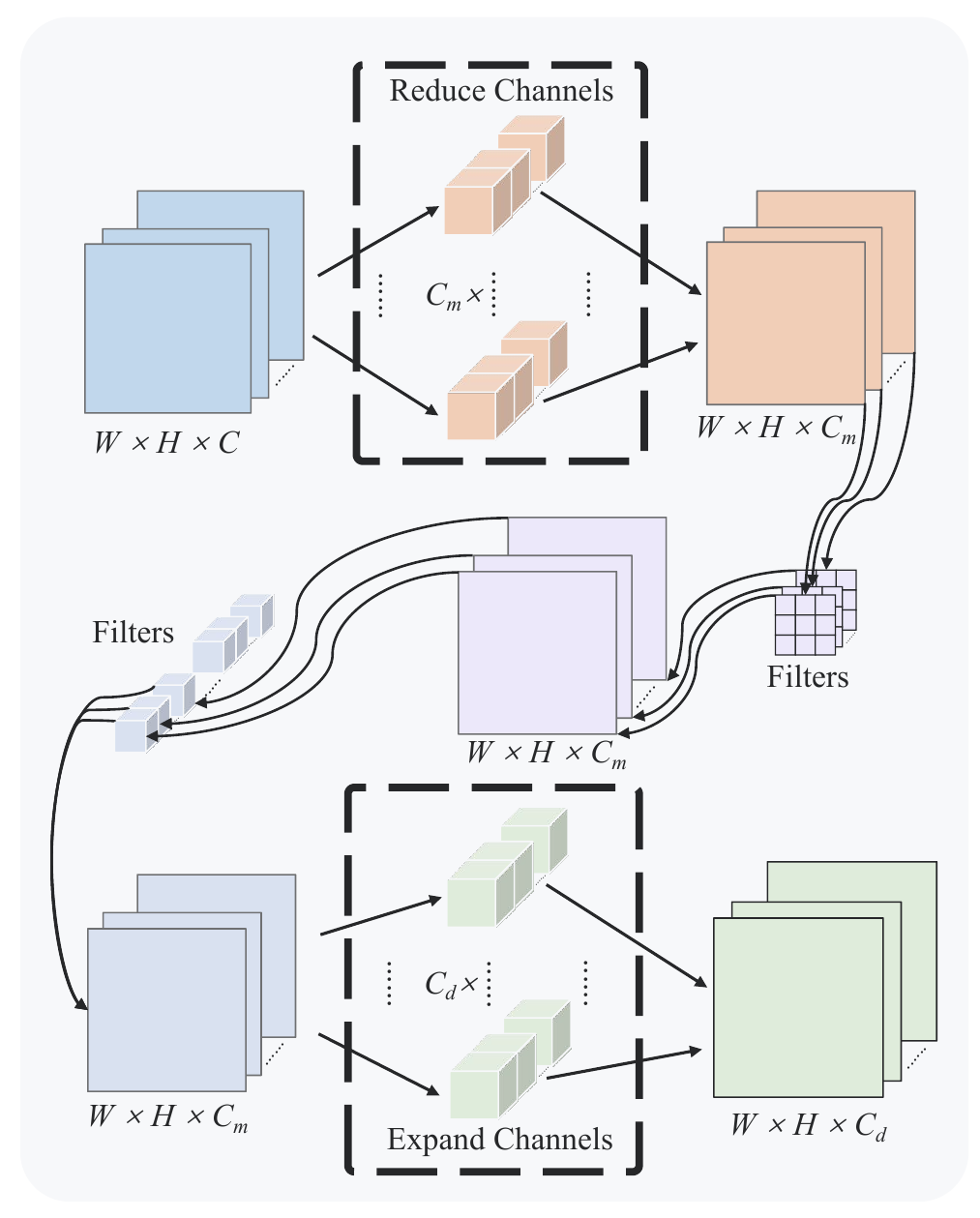}
  \caption{Illustration of LRDS convolution block. After the number of channels of the input features is reduced, their local and spatial features are efficiently extracted by depth-wise convolution and point-wise convolution, and finally the number of channels is restored.}
  \label{fig:LRDS}
\end{figure}

As shown in Figure \ref{fig:LRDS}, our designed LRDS block adopts a bottleneck-like structure. It comprises a Reduce layer, a depthwise convolution layer, a pointwise convolution layer, and an Expand layer. The input feature map first passes through the Reduce layer, reducing the input channels to \(C_m\). It then goes through the depthwise convolution layer and the pointwise convolution layer, keeping the number of channels unchanged. Finally, the Expand layer increases the number of channels to \(C_d\), producing the final output feature map.

Low-rank approximation decomposes a high-dimensional matrix into low-dimensional matrices, reducing computation complexity. In CNNs, it is applied to convolutional kernel responses, assuming these lie in a low-rank subspace.

Specifically, in the convolutional layer, the filter is assumed to have a spatial size of $k$ and a number of input channels of $c$. In computing the convolutional response, the filter is applied to a sub-tensor of the input tensor of shape $k \times k \times c$. Let this sub-tensor be denoted as $t \in \mathbb{R}^{k^2 c + 1}$, where a bias term is appended at the end. Consequently, the response at a position in the output, $y \in \mathbb{R}^e$, can be expressed as follows:

\vspace{-0.3cm}
\begin{equation}
\label{eq:1}
y = Wt
\end{equation}
where $e$ represents the number of filters and $W$ is an $e \times (k^2 c + 1)$ matrix. Assuming the vector $y$ lies in a low-rank subspace, it can be expressed as $y = U(y - y_1) + y_1$, where $U$ is an $e \times e$ matrix of rank $e_0$, with $e_0 < e$. Here, $y_1$ denotes the mean vector of the response. Therefore, the response can be formulated as

\begin{equation}
\label{eq:2}
y = UWt + b
\end{equation}
where $b = y_1 - Uy_1$ is the new bias. Given that the rank of $U$ is $e_0$, it can be decomposed into two $e \times e_0$ matrices $J$ and $K$ such that $U = JK^\top$. Let $W_1 = K^\top W$ denote an $e_0 \times (k^2 c + 1)$ matrix, which corresponds to a set of $e_0$ filters. Thus, the response can be rewritten as follows:

\begin{equation}
\label{eq:3}
y = JW_1 t + b
\end{equation}

The computational complexity of using Equation (\ref{eq:3}) is $O(e_0 k^2 c) + O(ee_0)$, whereas approximating $y$ in the low-rank subspace using Equation (\ref{eq:1}) has a computational complexity of $O(e k^2 c)$. Given that $O(ee_0) \ll O(e_0 k^2 c)$, the complexity of using Equation (\ref{eq:3}) is $\frac{e_0}{e}$ times that of using Equation (\ref{eq:1}).

Low-rank approximation theory significantly reduces the parameters and computational load of convolutional layers. To efficiently capture spatial information within the low-rank space for finer perception of detailed regions, the LRDS module processes features by embedding depthwise convolution and pointwise convolution directly into the low-rank space. Consequently, the MFE is constructed by replacing all standard convolution operations in the ResNet50 backbone with LRDS convolution blocks.

\subsection{Lightweight Long-range Dependency Extractor}

\begin{figure}[!t]
  \centering
  \includegraphics[width=0.7\textwidth]{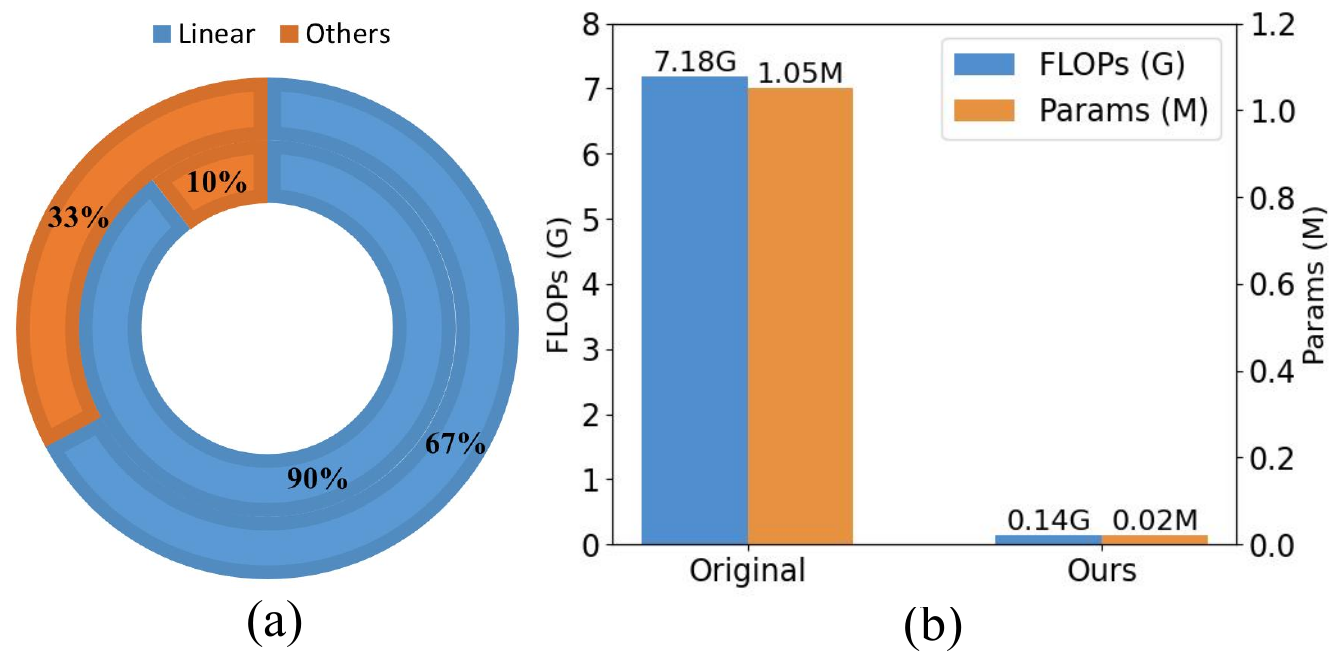}
  \caption{Analysis of Computational Resources Required for Linear Operations. (a) shows the percentage of FLOPs and Params in the LDE module that use Original linear operations, with the inner circle showing the percentage that use FLOPs and the outer circle showing the percentage of Params. (b) Shows the comparison of FLOPs and Params for LDE before and after using LR Linear.}
  \label{fig:Linear_Ana}
  \vspace{-0.3cm}
\end{figure}

When extracting long-range dependencies of crack pixels, the deformable attention mechanism adapts well to irregular structures. Specifically, given an input $x \in \mathbb{R}^{C \times H \times W}$, let $q$ denote a query element with query feature $z_q$ and reference point coordinates $(p_{qx}, p_{qy})$, the deformable attention is calculated as:

\vspace{-0.3cm}
\begin{equation}
\centering
\label{eq:multi-scale deformable attention}
\begin{split}
& \text{MSDeformAttn}(z_q, \hat{p}_q, \{x_c\}_{c=1}^C) 
= \\
& \sum_{h=1}^{H} W_h \Bigg[ \sum_{c=1}^{C} \sum_{t=1}^{T} A_{hcqt} \cdot 
 W'_h x_c(\phi_c(\hat{p}_q) + \Delta p_{hcqt}) \Bigg]
\end{split}
\end{equation}
Where $h$ represents the attention head, $c$ represents the input feature layer, and $t$ represents the sampling key point. $\Delta p_{hcqt}$ and $A_{hcqt}$ represent the offset and attention weight of the $t$-th sampling point in the $c$-th feature map and $h$-th attention head, respectively. The normalization is such that $\sum_{c=1}^{C} \sum_{t=1}^{T} A_{hcqt} = 1$. Here, normalized coordinates $\hat{p} \in [0,1]$ are used to represent the position of a reference point in the normalized feature map $\phi_{c}(\hat{p}_q)$ rescales the normalized coordinates onto the $c$-th layer feature map.

As shown in Figure \ref{fig:Linear_Ana}(a), Linear operations account for 90\% of the FLOPs and 67\% of the Params in the LDE module. To reduce the computational resources required by the LDE, a linear transformation method (LR Linear) based on low-rank approximation is proposed. In traditional linear layers, the weight matrix connecting the input and output is typically a high-rank matrix, representing a complete connection between input and output channels. By applying low-rank approximation, the weight matrix is decomposed into two smaller matrices, effectively reducing the number of parameters.

Specifically, the weight matrix \( E \in \mathbb{R}^{r_{\text{out}} \times r_{\text{in}}} \), where \( r_{\text{out}} \) is the output feature dimension and \( r_{\text{in}} \) is the input feature dimension, is approximated by two smaller matrices \( A \in \mathbb{R}^{r_{\text{out}} \times r} \) and \( B \in \mathbb{R}^{r_{\text{in}} \times r} \), where \( r \) is a small rank, much smaller than \( r_{\text{out}} \) and \( r_{\text{in}} \). This decomposition breaks the original full connection into two smaller linear transformations:

\vspace{-0.3cm}
\begin{equation}
E \approx E' = A B^T
\end{equation}

This transformation is performed through two linear layers. First, a linear transformation $B^\top$ is applied to the input $x \in \mathbb{R}^{r_{\text{in}}}$, resulting in an intermediate representation $z \in \mathbb{R}^{r}$:

\vspace{-0.3cm}
\begin{equation}
z = B^T x
\end{equation}

Subsequently, another linear transformation $A$ is applied to $z$, resulting in the output $y \in \mathbb{R}^{r_{\text{out}}}$:

\vspace{-0.3cm}
\begin{equation}
y = A z
\end{equation}

As \( r \ll r_{\text{out}}, r_{\text{in}} \), the low-rank approximation reduces the original matrix to two smaller matrices, significantly reducing the computational and storage requirements. This leads to a reduction in both the computational complexity and the number of parameters, thus improving training and inference efficiency. As shown in Figure \ref{fig:Linear_Ana}(b), replacing all Linear operations in the LDE with LR Linear reduced the parameters and computational load from 7.18G and 1.05M to 0.14G and 0.02M.

The structure of LDE is shown in Figure \ref{fig:Network_Arch}. LDE facilitates long-range dependency modeling through the interaction of multi-scale features between the encoder and decoder. Initially, the encoder inputs a multi-scale feature set $\{X_c\}_{c=1}^C$ (where $C$ represents the number of feature levels). After undergoing layer flattening and positional encoding enhancement, the encoder extracts the feature sequence $S \in \mathbb{R}^{B \times L \times d}$ (where $L$ denotes the sequence length). The encoder then combines LR Linear with lightweight deformable attention mechanisms to capture long-range dependencies:

\begin{equation}
Z = \text{MSDeformAttn}(S, P, S)
\end{equation}
where $P \in \mathbb{R}^{B \times L \times 2}$ represent the normalized reference coordinates. Subsequently, $Z$ is processed through a feedforward network (FFN) that combines LR Linear to refine the features:

\begin{equation}
M = \text{LR\_Linear} (\text{GELU}(\text{LR\_Linear}(Z)))
\end{equation}
where the output memory matrix $M \in \mathbb{R}^{B \times L \times d}$ is used as the input to the decoder, encoding multi-scale local dependencies.

The decoder processes the input with a learnable query vector \( Q \in \mathbb{R}^{N \times d} \) and the initial reference point \( H = \text{Sigmoid}(T Q) \), where \( T \in \mathbb{R}^{d \times 2} \) is the projection matrix. First, the reference points \( H \in \mathbb{R}^{N \times 2} \) are mapped to the encoded memory \( M \) to aggregate features:

\begin{equation}
Y_0 = \text{MSDeformAttn}(Q, H, M)
\end{equation}

Then, it passes through the FFN to obtain the crack pixel sequence enriched with irregular long-range dependencies:
\begin{equation}
Y = \text{LR\_Linear} (\text{GELU}(\text{LR\_Linear}(Y_0)))
\end{equation}
where Y is the output sequence of the LDE enriched with irregular long-range dependencies between pixels.

\subsection{Staircase Cascaded Fusion Module}

\begin{figure}[!t]
  \centering
  \includegraphics[width=0.6\textwidth]{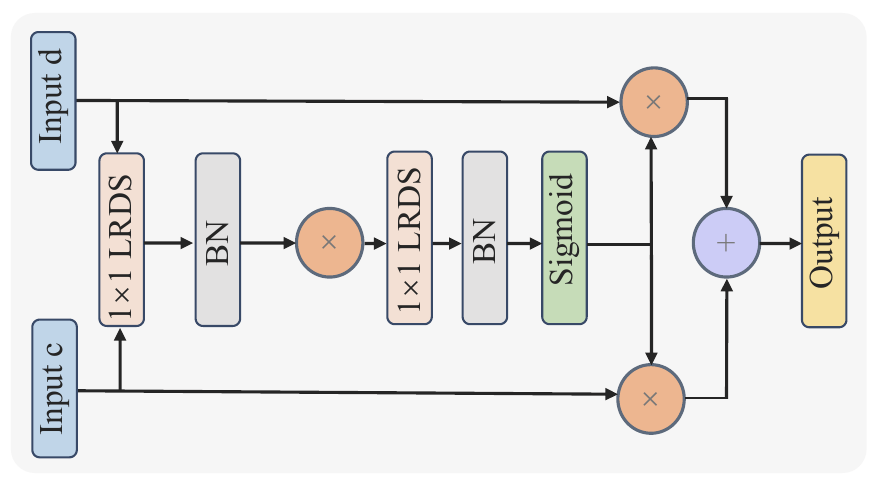}
  \caption{Illustration of the PAF. It effectively reduces required computing resources by embedding LRDS.}
  \label{fig:PAF}
  \vspace{-0.4cm}
\end{figure}

The design of our SCFM structure is shown in Figure \ref{fig:Network_Arch}. As a segmentation head, to enhance useful semantic features in the MFE and LDE branches and ensure effective semantic information is not buried, this paper proposed the lightweight Pixel Attention Guided Fusion (PAF) module. The structure of PAF is shown in Figure \ref{fig:PAF}. We denote the pixels in the feature maps of the two branches as $\vec{v_c}$ and $\vec{v_d}$, respectively. The output can be expressed as:

\vspace{-0.3cm}
\begin{equation}
\label{eq:PAF1}
\sigma = \text{Sigmoid}(f_c(\vec{v_c}) \cdot f_d(\vec{v_d}))
\end{equation}
where $f_c$ and $f_d$ denote pointwise convolution and batch normalization operations, respectively. Thus, the output of PAF can be represented as: 

\vspace{-0.3cm}
\begin{equation}
\label{eq:PAF2}
\text{Output} = \sigma \vec{v_c} + (1-\sigma) \vec{v_d}
\end{equation}
where $\sigma$ indicates the probability that the two pixels belong to the same category. If $\sigma$ is high, the model will trust $\vec{v_c}$ more, indicating that its semantic features are more accurate, and vice versa.

To reduce computational resource usage, all convolution operations in this module utilize LRDS blocks. As seen in Figure \ref{fig:FLOPS_Params_Ana}, using original convolution operations in the SCFM results in FLOPs and Params reaching 73.36G and 9.71M, respectively. Thus, lightweighting these operations is crucial. Replacing all convolution operations with LRDS blocks reduces FLOPs and Params by 92.87\% and 97.22\%, respectively, demonstrating the effectiveness of LRDS in network lightweighting.

SCFM is divided into four stages, each with two fusion branches: the concat branch and the PAF branch. The concat operation retains complete information and diverse feature representations through direct concatenation, while the PAF module selectively fuses features through a pixel-guided attention mechanism, avoiding issues like buried feature information, information conflict, and redundancy. The processes of SCFM can be given by the following equation:


\begin{equation}
\begin{split}
& F_i^{\text{O}} = C_{\text{i}} \left(F_i^{\text{MFE}}, F_{i-1}^{\text{O}} \right) \\
& + \mu \left( P_i \left( F_i^{\text{MFE}}, F_{i-1}^{\text{O}} \right) \right) \odot P_i \left( F_i^{\text{MFE}}, F_{i-1}^{\text{O}} \right)
\end{split}
\end{equation}
Where \( F_i^{\text{O}} \) represents the output features of the \( i \)-th SCFM layer (\( i \) ranges from 0 to 3), \( F_i^{\text{MFE}} \) represents the \( i \)-th layer MFE features, \( C_{\text{i}} \) is an operation chain: Concat followed by LRDS, ReLU, and GN, \( P_i \) is a PAF operation, \( \mu(\cdot) \) is the sigmoid function generating spatial attention weights, and \( \odot \) represents the Hadamard product. Specifically, when \( i = 0 \), \( F_{i-1}^{\text{O}} \) is defined as the output of LDE, i.e., \( F_{-1}^{\text{O}} = \text{LDE}_{\text{output}} \), ensuring an initial output feature is provided at the beginning of the computation.

By perceiving the feature maps from the two branches, both local patterns of crack pixels and long-range dependencies can be comprehensively addressed. To fully utilize the outputs of MFE and LDE and improve segmentation performance, SCFM utilized the property that PAF can process multiple feature maps simultaneously by merging the outputs of all the preceding phases and the associated MFE feature maps from the second phase onwards. The processing enables SCFM to more comprehensively dissect and process the local patterns and long-range dependencies in the different stages of the feature map to achieve a finer and more continuous segmentation map. Furthermore, the fine-grained features extracted with PAF can be perceived more finely through the dynamic weighting mechanism to achieve a clearer segmentation map.

\subsection{Loss Function}

The discrepancy between predictions and actual values is quantified by the loss function; a smaller value for this loss suggests superior model performance. Previously, Binary Cross-Entropy (BCE) \cite{li2024rediscovering} was often used, but it may not effectively distinguish the minority crack pixels in the image. Combining BCE with Dice loss \cite{li2024cnn} addresses this issue. BCE improves the probability distribution for each pixel, while Dice loss maximizes the overlap between predictions and true labels, enhancing robustness against imbalanced data.

Our loss $L$ can be calculated using Equation (\ref{eq:our_loss}), which combines the above two loss functions for joint optimization.

\vspace{-0.3cm}
\begin{equation}
\label{eq:our_loss}
L = 0.75 \cdot L_{B} + 0.25 \cdot L_{D}
\end{equation}
where $\mathit{L_{B}}$ represents the BCE loss function and $\mathit{L_{D}}$ represents the Dice loss function.

In this section, the paper analyzes the differences between the TUT dataset and other public datasets, describes the experimental setup, and compares various methods and metrics. Subsequently, the results and model complexity are analyzed, and ablation experiments are performed.

\section{Experimental Results and Discussion}

\subsection{Dataset Description}

\begin{table*}[!t]
    \centering
    \caption{Comparison of Different Crack Datasets}
    \begin{tabular}{cccccccc}
    \toprule
        No. & Datasets & Num & Resolution & Crack Pixels & Scenarios \\ \hline
        1 & CFD \cite{shi2016automatic} & 118 & 480 $\times$ 320 & 1.60\% & 1 \\ 
        2 & GAPS509 \cite{eisenbach2017get} & 509 & 540 $\times$ 640 & 1.19\% & 1 \\ 
        3 & DeepCrack \cite{liu2019deepcrack} & 537 & 544  $\times$  384 & 3.54\% & 2 \\ 
        4 & Crack500 \cite{yang2019feature} & 3368 & 640  $\times$  360 & 6.01\% & 2 \\ 
        5 & CrackMap \cite{katsamenis2023few} & 120 & 256  $\times$  256 & 4.10\% & 1 \\ 
        6 & AEL \cite{yang2019feature} & 58 & 768  $\times$  512 & 0.67\% & 1 \\ 
        7 & CrackTree206 \cite{zou2012cracktree} & 206 & 800  $\times$  600 & 0.32\% & 1 \\ 
        8 & LSCD \cite{yao2024cracknex} & 143 & 400  $\times$  400 & 3.18\% & 2 \\ 
        9 & CrackSC \cite{guo2023pavement} & 197 & 320  $\times$  480 & 2.02\% & 2 \\ \hline

        10 & \textbf{TUT (Ours)} & 1408 & 640 $\times$  640 & 3.16\% & 8 \\ 
    \bottomrule
    \end{tabular}
    \label{tab:dataset}
    \vspace{-0.2cm}
\end{table*}

\begin{figure*}[!t]
  \centering
  \includegraphics[width=0.98\textwidth]{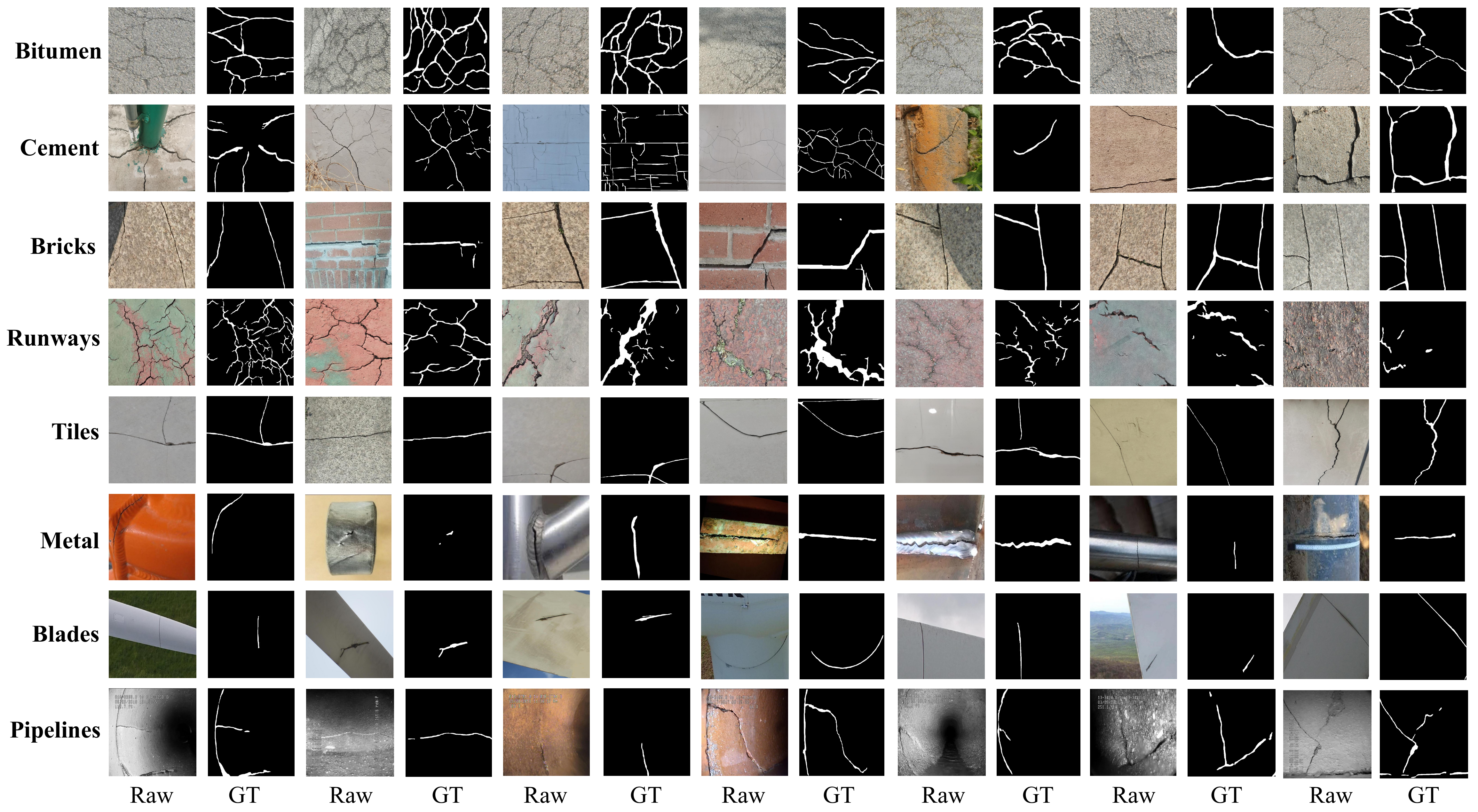}
  \caption{Illustration of the images of each scenarios in the TUT dataset.}
  \label{fig:TUT_Dataset}
  \vspace{-0.5cm}
\end{figure*}

The previous works has used dataset scenarios that are relatively simple, such as cracks in cement or bitumen pavements only. These datasets have clear backgrounds and minimal interference, which cannot comprehensively evaluate the performance of models. To tackle this, this paper created the TUT dataset to enable better segmentation across different scenarios. Detailed information and comparisons with other datasets are list in Table \ref{tab:dataset}.

The TUT dataset comprises 1270 crack images captured with our mobile phones, along with a small number of 138 images sourced from the Internet. TUT dataset contains 900 images of bitumen and cement with irregular crack morphology but flat backgrounds, 142 images of bricks and tiles with regular crack morphology, 228 images of plastic runways with extremely complex cracks, and a total of 138 images of metals, generator blades, and underground pipelines with complex lighting and background disturbances. To ensure reliability in model training and testing, all images were manually annotated by several researchers to generate binary labels, and the best-annotated images were selected as the final labeled images. Unlike other datasets with clear backgrounds, the TUT dataset images have complex, noisy backgrounds and cracks of intricate shapes. This evaluates the model's capability to segment cracks in challenging conditions.

As shown in Table \ref{tab:dataset}, our TUT dataset contains eight image scenarios, far more than the 1-2 scenarios of other public datasets, this allows for a more comprehensive evaluation of the model's performance in detecting cracks under various scenarios. Figure \ref{fig:TUT_Dataset} shows crack images in different scenarios and their binary labeled images.  The crack pixel ratio in the TUT dataset is 3.16\%, which is moderate. This ensures the model learns useful information without being hindered by too few crack pixels and prevents the model from overly relying on the crack pixel ratio when there are too many crack pixels.

As seen in Figure \ref{fig:TUT_Dataset}, bitumen, cement, bricks, and plastic runways have extremely complex cracks due to their unique material properties. Particularly in plastic runways, both coarse and fine cracks are present, testing the model's generalization to different types of cracks. In images of tiles, metal materials, wind turbine blades, and underground pipelines, the crack shapes are relatively simple, but the backgrounds are very complex with various external interferences such as low light, irrelevant areas, and surface characteristics. This can easily result in false detections, thereby testing the model's ability to accurately extract crack areas from complex environments.

\subsection{Implementation Details}

\textit{1) Experimental Settings:} The CrackSCF network is built on the PyTorch v2.1.1 deep learning framework and trained on an Ubuntu 20.04 server, featuring two Intel(R) Xeon(R) Platinum 8336C CPUs and eight Nvidia GeForce 4090 GPUs, each with 24GB VRAM. For training, the AdamW optimizer is utilized with an initial learning rate set at 0.0005, weight decay of 0.0001, batch size of one, and a total of 60 training epochs. The learning rate is reduced to one-tenth after 30 epochs as part of a decay strategy. The best-performing model parameters on the validation set are selected for testing. We compared the CrackSCF network with other SOTA methods on six datasets: CFD \cite{shi2016automatic}, GAPS509 \cite{eisenbach2017get}, DeepCrack \cite{liu2019deepcrack}, Crack500 \cite{yang2019feature}, CrackMap \cite{katsamenis2023few}, and ours TUT dataset.

\textit{2) Comparison Methods:} In evaluating model performance, we compared the CrackSCF network with 12 classic segmentation methods and the latest SOTA methods on different datasets. These methods include HED \cite{xie2015holistically}, UNet++ \cite{zhou2018unet++}, Deeplabv3+ \cite{chen2018encoder}, AttuNet \cite{oktay2018attention}, RCF \cite{liu2017richer}, RIND \cite{pu2021rindnet}, UCTNet \cite{wang2022uctransnet}, SFIAN \cite{cheng2023selective}, CT-crackseg \cite{tao2023convolutional}, DTrCNet \cite{xiang2023crack}, Crackmer \cite{wang2024dual} and Simcrack \cite{jaziri2024designing}.

\begin{figure*}[!t]
  \centering
  \includegraphics[width=\textwidth]{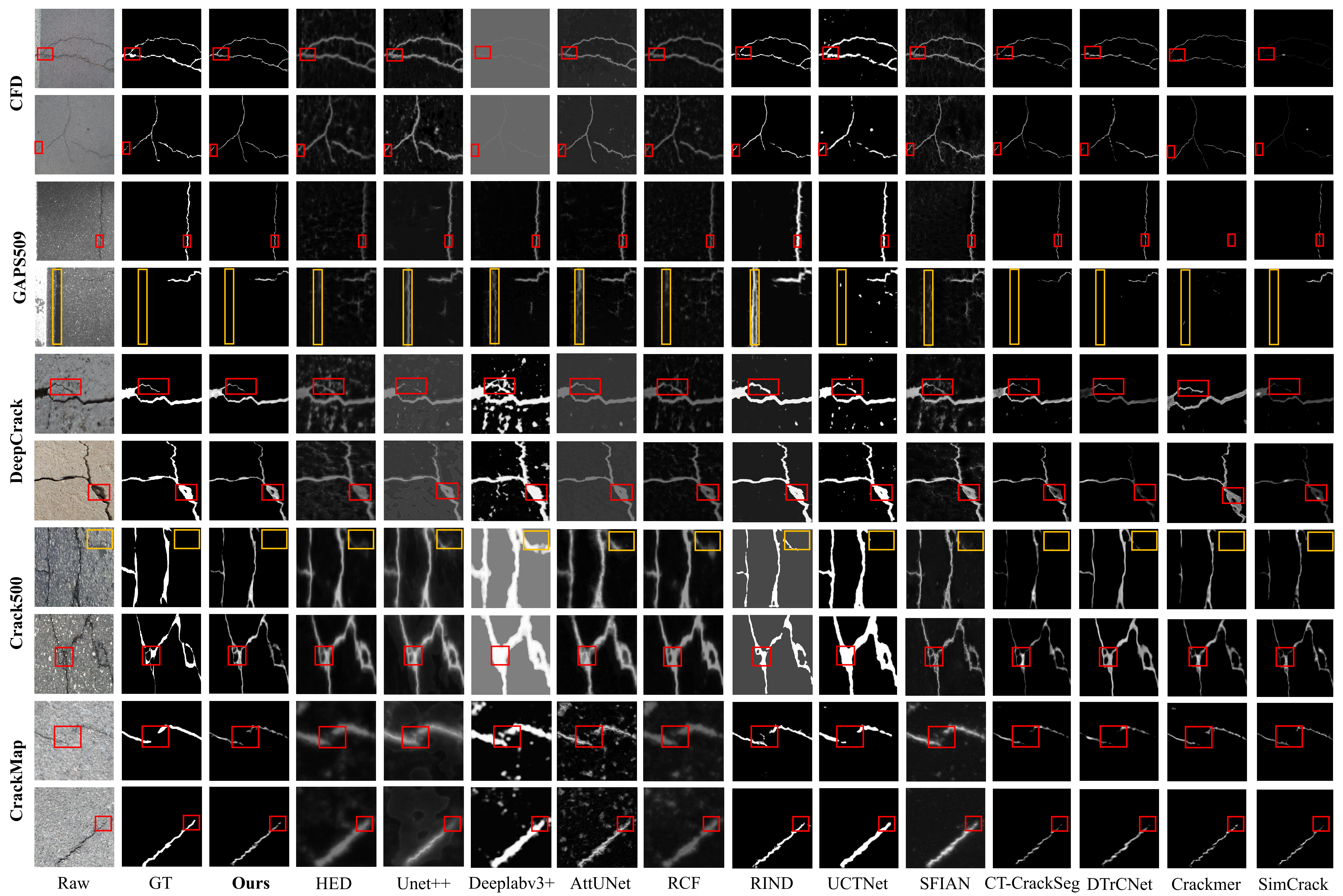}
  \caption{Visualization results of CrackSCF and other methods on public datasets. Key areas are highlighted with red boxes to indicate easily missed crack pixels and yellow boxes to indicate easily misdetected non-crack areas.}
  \label{fig:Public dataset result}
  \vspace{-0.5cm}
\end{figure*}

\textit{3) Evaluation Metric:} To quantitatively evaluate the performance of our proposed crack segmentation model, we used six main metrics: Precision (P), Recall (R), F1 Score ($F1 = \frac{2PR}{P + R}$), Mean Intersection over Union (mIoU), Optimal Dataset Scale (ODS), and Optimal Image Scale (OIS). Mean Intersection over Union (mIoU) measures the average ratio of the intersection and union of the predicted results and the true results. The calculation formula for mIoU is:
\begin{equation}
\label{eq:mIoU}
mIoU = \frac{1}{i + 1} \sum_{a=0}^{i} \frac{p_{aa}}{\sum_{b=0}^{i} p_{ab} + \sum_{b=0}^{i} p_{ba} - p_{aa}}
\end{equation}
where $i$ is the number of classes, $a$ denotes the true value, $b$ denotes the predicted value, and $p_{ab}$ represents the number of pixels predicted as $a$ for $b$. Given that our crack segmentation task involves binary classification, we set $i = 1$.

Optimal Dataset Scale (ODS) indicates the optimal segmentation performance metric obtained when a fixed threshold $t$ is selected over the entire dataset. It is defined as:

\begin{equation}
\label{eq:ODS}
ODS = \max_{t} \frac{2 \cdot P_t \cdot R_t}{P_t + R_t}
\end{equation}

Optimal Image Scale (OIS) indicates the optimal segmentation performance metric obtained when an optimal threshold $t$ is selected for each image individually. It is defined as:

\begin{equation}
\label{eq:OIS}
OIS = \frac{1}{N} \sum_{a=1}^{N} \max_{t} \frac{2 \cdot P_{t,a} \cdot R_{t,a}}{P_{t,a} + R_{t,a}}
\end{equation}
where $N$ represents the total number of images.

\subsection{Comparison with the SOTA methods}

Our proposed method, CrackSCF, was compared with 12 other methods across five public datasets and our own TUT dataset.

\begin{table}[!t]
  \centering
  \caption{Comparison results on the CFD dataset. Best results are bolded and second best results are underlined.}
  \setlength{\tabcolsep}{1.5mm}{
  \begin{tabular}{lcccccc}
    \toprule
    Methods & ODS & OIS & P & R & F1 & mIoU \\
    \midrule
    HED & 0.6276 & 0.6494 & 0.5467 & 0.7374 & 0.6279 & 0.7240 \\
    UNet++ & 0.6586 & 0.6836 & 0.5995 & 0.7608 & 0.6706 & 0.7432 \\
    DeeplabV3+ & 0.6910 & 0.7084 & 0.6264 & \underline{0.7730 } & 0.6920 & 0.7709 \\
    AttuNet & 0.6710 & 0.6870 & 0.6240 & 0.7220 & 0.6694 & 0.7576 \\
    RCF & 0.6434 & 0.6624 & 0.5792 & 0.7348 & 0.6478 & 0.7343 \\
    RIND & 0.6858 & 0.7021 & 0.6456 & 0.7534 & 0.6954 & 0.7592 \\
    UCTNet & 0.6587 & 0.7006 & 0.6701 & 0.7205 & 0.6944 & 0.7479 \\
    SFIAN & 0.7040 & 0.7159 & 0.6657 & 0.7512 & 0.7058 & 0.7684 \\
    CT-crackseg & 0.7122 & 0.7179 & \underline{0.6896 } & 0.7438 & 0.7157 & 0.7738 \\
    DTrCNet & 0.7079 & 0.7123 & 0.6774 & 0.7486 & 0.7112 & 0.7710 \\
    Crackmer & 0.6751 & 0.6754 & 0.6370 & 0.7237 & 0.6776 & 0.7503 \\
    SimCrack & \underline{0.7152 } & \underline{0.7204 } & 0.6883 & 0.7557 & \underline{0.7204 } & \underline{0.7756 } \\
    \textbf{Ours} & \textbf{0.7335 } & \textbf{0.7376 } & \textbf{0.6967 } & \textbf{0.7848 } & \textbf{0.7383 } & \textbf{0.7864 } \\
    \bottomrule
  \end{tabular}}
  \label{tab:CFD}
\end{table}

\textit{1) Results on the CFD dataset:} The crack areas in the CFD dataset are very slender, testing the model's sensitivity to fine-grained crack pixels. As shown in Table \ref{tab:CFD}, our proposed CrackSCF attained the best results across all metrics for the CFD dataset. Specifically, on ODS, OIS, P, R, F1, and mIoU, it outperformed the second-best methods by 0.48\%, 0.34\%, 0.14\%, 0.35\%, 0.31\%, and 0.30\%, respectively. Notably, SimCrack \cite{jaziri2024designing} achieved the second-best performance on all metrics except for P and R, only behind our method. However, as seen in Figure \ref{fig:Public dataset result}, its actual segmentation performance is not satisfactory, while our method achieves high-quality segmentation results on various fine-grained crack pixels. Other methods, such as UCTNet \cite{wang2022uctransnet} and DTrCNet \cite{xiang2023crack}, tend to misdetect background noise or miss key detail areas of crack pixels.

\begin{table}[!t]
  \centering
  \caption{Comparison results on the GAPS509 dataset. Best results are bolded and second best results are underlined.}
  \setlength{\tabcolsep}{1.5mm}{
  \begin{tabular}{lcccccc}
    \toprule
    Methods & ODS & OIS & P & R & F1 & mIoU \\
    \midrule
    HED & 0.4014 & 0.4509 & 0.4300 & 0.4624 & 0.4456 & 0.6282 \\
    UNet++ & 0.6255 & 0.6484 & 0.6160 & 0.6684 & 0.6411 & 0.7300 \\
    DeeplabV3+ & 0.6160 & 0.6425 & 0.6050 & 0.6669 & 0.6345 & \underline{0.7378 } \\
    AttuNet & 0.5955 & 0.6174 & 0.5924 & 0.6519 & 0.6208 & 0.7247 \\
    RCF & 0.4532 & 0.4817 & 0.4420 & 0.5428 & 0.4872 & 0.6472 \\
    RIND & 0.5505 & 0.6023 & 0.4959 & 0.6494 & 0.5623 & 0.6909 \\
    UCTNet & 0.5835 & 0.6188 & 0.5709 & 0.6752 & 0.6187 & 0.7107 \\
    SFIAN & 0.6087 & 0.6320 & 0.6061 & 0.6451 & 0.6250 & 0.7198 \\
    CT-crackseg & 0.6029 & 0.6075 & 0.5693 & 0.6804 & 0.6199 & 0.7165 \\
    DTrCNet & 0.6211 & 0.6276 & \underline{0.6277 } & 0.6592 & 0.6431 & 0.7272 \\
    Crackmer & 0.4593 & 0.4642 & 0.6144 & 0.4663 & 0.5302 & 0.6581 \\
    SimCrack & \underline{0.6408 } & \underline{0.6471 } & 0.6171 & \textbf{0.6997 } & \underline{0.6558 } & 0.7368 \\
    \textbf{Ours} & \textbf{0.6492 } & \textbf{0.6552 } & \textbf{0.6459 } & \underline{0.6812 } & \textbf{0.6631 } & \textbf{0.7411 } \\
    \bottomrule
  \end{tabular}}
  \label{tab:GAPS509}
\end{table}

\textit{2) Results on the GAPS509 dataset:} Unlike the CFD dataset, the GAPS509 dataset contains not only finer cracks but also some irrelevant background interference, testing the model's ability to extract crack areas under high interference conditions. As shown in Table \ref{tab:GAPS509}, our method achieved the highest performance on all metrics except R. Although SimCrack \cite{jaziri2024designing} achieved 0.6997 on the R metric, leading our method by about 2.72\%, its P metric was 0.6171, far lower than our method. The highest F1 score obtained by our method indicates a good balance between P and R. Specifically, our method outperformed the second-best methods on ODS, OIS, P, F1, and mIoU by 1.31\%, 1.25\%, 2.90\%, 1.11\%, and 0.45\%, respectively. Additionally, as seen in Figure \ref{fig:Public dataset result}, our method almost perfectly avoids irrelevant background noise, while most other methods falsely detect irrelevant backgrounds and perform worse in segmenting crack pixel areas compared to our method.

\begin{table}[!t]
  \centering
  \caption{Comparison results on the DeepCrack dataset. Best results are bolded and second best results are underlined.}
  \setlength{\tabcolsep}{1.5mm}{
  \begin{tabular}{lcccccc}
    \toprule
    Methods & ODS & OIS & P & R & F1 & mIoU \\
    \midrule
    HED & 0.7552 & 0.8237 & 0.7231 & 0.7678 & 0.7448 & 0.8007 \\
    UNet++ & 0.8419 & 0.8893 & 0.8474 & 0.8533 & 0.8503 & 0.8621 \\
    DeeplabV3+ & 0.7727 & 0.8354 & 0.7363 & 0.8323 & 0.7813 & 0.8565 \\
    AttuNet & 0.8153 & 0.8379 & 0.8066 & 0.8069 & 0.8068 & 0.8578 \\
    RCF & 0.7666 & 0.8376 & 0.7423 & 0.7611 & 0.7516 & 0.8110 \\
    RIND & 0.8087 & 0.8267 & 0.7896 & \underline{0.8920 } & 0.8377 & 0.8391 \\
    UCTNet & 0.8357 & 0.8579 & 0.8217 & 0.8857 & 0.8525 & 0.8564 \\
    SFIAN & 0.8616 & \underline{0.8928 } & 0.8549 & 0.8692 & 0.8620 & 0.8776 \\
    CT-crackseg & \underline{0.8819 } & 0.8904 & \underline{0.9011 } & 0.8895 & \underline{0.8952 } & \underline{0.8925 } \\
    DTrCNet & 0.8473 & 0.8512 & 0.8905 & 0.8251 & 0.8566 & 0.8661 \\
    Crackmer & 0.8712 & 0.8785 & 0.8946 & 0.8783 & 0.8864 & 0.8844 \\
    SimCrack & 0.8570 & 0.8722 & 0.8984 & 0.8549 & 0.8761 & 0.8744 \\
    \textbf{Ours} & \textbf{0.8914 } & \textbf{0.8963 } & \textbf{0.9147 } & \textbf{0.9013 } & \textbf{0.9079 } & \textbf{0.9001 } \\
    \bottomrule
  \end{tabular}}
  \label{tab:DeepCrack}
\end{table}

\textit{3) Results on the DeepCrack dataset:} The cracks in the DeepCrack dataset are often relatively wide, testing the model's ability to extract a large range of crack pixels. As shown in Table \ref{tab:DeepCrack}, our method achieved the highest results on all metrics for the DeepCrack dataset. Specifically, our method outperformed the second-best methods on ODS, OIS, P, R, F1, and mIoU by 1.08\%, 0.39\%, 1.51\%, 1.04\%, 1.42\%, and 0.85\%, respectively. Notably, CT-crackseg also achieved good results on this dataset, but as seen in Figure \ref{fig:Public dataset result}, its segmentation performance on detailed cracks is not as good as our method, indicating that our method can effectively handle crack details.

\textit{4) Results on the Crack500 dataset:} Similar to the DeepCrack, the Crack500 dataset also has relatively wide crack areas, but its structure is complex, posing higher requirements on model performance. As shown in Table \ref{tab:Crack500}, our method achieved the best results on five metrics for the Crack500 dataset, except for R. It outperformed the second-best methods on ODS, OIS, P, F1, and mIoU by 1.11\%, 0.07\%, 1.69\%, 0.47\%, and 0.64\%, respectively. Although DTrCNet \cite{xiang2023crack} leads on the R metric, our method surpasses it on the P metric by 10.51\%. The highest F1 score achieved by our method indicates a good balance between P and R. Additionally, as seen in Figure \ref{fig:Public dataset result}, our method does not produce false detections in irrelevant background areas, whereas other methods either produce false detections or lack detailed segmentation in crack areas.

\begin{table}[!t]
  \centering
  \caption{Comparison results on the Crack500 dataset. Best results are bolded and second best results are underlined.}
  \setlength{\tabcolsep}{1.5mm}{
  \begin{tabular}{lcccccc}
    \toprule
    Methods & ODS & OIS & P & R & F1 & mIoU \\
    \midrule
    HED & 0.6191 & 0.6537 & 0.6150 & 0.7281 & 0.6668 & 0.7140 \\
    UNet++ & 0.6965 & 0.7274 & 0.6989 & 0.7805 & 0.7374 & 0.7605 \\
    DeeplabV3+ & 0.6943 & 0.7211 & 0.6970 & 0.7569 & 0.7257 & 0.7746 \\
    AttuNet & 0.6586 & 0.6890 & 0.6490 & 0.7608 & 0.7005 & 0.7543 \\
    RCF & 0.6054 & 0.6361 & 0.6037 & 0.7136 & 0.6540 & 0.7068 \\
    RIND & 0.6469 & 0.6483 & 0.6998 & 0.7245 & 0.7119 & 0.7381 \\
    UCTNet & \underline{0.7147 } & 0.7265 & 0.6887 & 0.7843 & 0.7302 & 0.7674 \\
    SFIAN & 0.6977 & \underline{0.7348 } & 0.6983 & 0.7742 & 0.7343 & 0.7604 \\
    CT-crackseg & 0.6941 & 0.7059 & 0.6940 & 0.7748 & 0.7322 & 0.7591 \\
    DTrCNet & 0.7012 & 0.7241 & 0.6527 & \textbf{0.8280 } & 0.7357 & 0.7627 \\
    Crackmer & 0.6933 & 0.7097 & 0.6985 & 0.7572 & 0.7267 & 0.7591 \\
    SimCrack & 0.7127 & 0.7308 & \underline{0.7093 } & 0.7984 & \underline{0.7516 } & \underline{0.7715 } \\
    \textbf{Ours} & \textbf{0.7226 } & \textbf{0.7353 } & \textbf{0.7213 } & \underline{0.7922 } & \textbf{0.7551 } & \textbf{0.7764 } \\
    \bottomrule
  \end{tabular}}
  \label{tab:Crack500}
  \vspace{-0.5cm}
\end{table}

\textit{5) Results on the CrackMap dataset:} Similar to the CFD dataset, the cracks in the CrackMap dataset are also relatively slender. As shown in Table \ref{tab:CrackMap}, our method leads all other methods on all metrics, outperforming the second-best methods on ODS, OIS, P, R, F1, and mIoU by 0.48\%, 0.34\%, 0.14\%, 0.35\%, 0.31\%, and 0.30\%, respectively. Notably, SimCrack \cite{jaziri2024designing} achieved the second-best results on this dataset, but as seen in Figure \ref{fig:Public dataset result}, SimCrack's segmentation performance on the crack tips is not satisfactory, whereas our method effectively identifies these pixels.

\begin{table}[!t]
  \centering
  \caption{Comparison results on the CrackMap dataset. Best results are bolded and second best results are underlined.}
  \setlength{\tabcolsep}{1.5mm}{
  \begin{tabular}{lcccccc}
    \toprule
    Methods & ODS & OIS & P & R & F1 & mIoU \\
    \midrule
    HED & 0.6022 & 0.6240 & 0.5281 & 0.7063 & 0.6043 & 0.7026 \\
    UNet++ & 0.7030 & 0.7222 & 0.6773 & 0.7209 & 0.6984 & 0.7631 \\
    DeeplabV3+ & 0.6617 & 0.6793 & 0.5853 & 0.7532 & 0.6587 & 0.7460 \\
    AttuNet & 0.6618 & 0.6773 & 0.6233 & 0.6970 & 0.6581 & 0.7454 \\
    RCF & 0.5552 & 0.5683 & 0.4590 & 0.7164 & 0.5595 & 0.6758 \\
    RIND & 0.6745 & 0.6943 & 0.6023 & 0.7586 & 0.6699 & 0.7425 \\
    UCTNet & 0.6104 & 0.6195 & 0.5653 & 0.6748 & 0.6152 & 0.7079 \\
    SFIAN & 0.7200 & 0.7465 & 0.6715 & 0.7668 & 0.7160 & 0.7748 \\
    CT-crackseg & 0.7289 & 0.7373 & 0.6911 & 0.7669 & 0.7270 & 0.7785 \\
    DTrCNet & 0.7328 & 0.7413 & 0.6912 & \underline{0.7681 } & 0.7276 & 0.7812 \\
    Crackmer & 0.7395 & 0.7437 & 0.7229 & 0.7467 & 0.7346 & 0.7860 \\
    SimCrack & \underline{0.7559 } & \underline{0.7625 } & \underline{0.7380 } & 0.7672 & \underline{0.7523 } & \underline{0.7963 } \\
    \textbf{Ours} & \textbf{0.7595 } & \textbf{0.7651 } & \textbf{0.7390 } & \textbf{0.7708 } & \textbf{0.7546 } & \textbf{0.7987 } \\
    \bottomrule
  \end{tabular}}
  \label{tab:CrackMap}
\end{table}

\begin{table}[!t]
  \centering
  \caption{Comparison results on the TUT dataset. Best results are bolded and second best results are underlined.}
  \setlength{\tabcolsep}{1.5mm}{
  \begin{tabular}{lcccccc}
    \toprule
    Methods & ODS & OIS & P & R & F1 & mIoU \\
    \midrule
    HED & 0.6960 & 0.7254 & 0.7080 & 0.7431 & 0.7251 & 0.7683 \\
    UNet++ & 0.7702 & 0.7948 & 0.7725 & 0.7974 & 0.7848 & 0.8160 \\
    DeeplabV3+ & 0.7504 & 0.7777 & 0.7155 & 0.8231 & 0.7655 & 0.8177 \\
    AttuNet & 0.7539 & 0.7795 & 0.7558 & 0.7945 & 0.7747 & 0.8204 \\
    RCF & 0.6821 & 0.7138 & 0.7106 & 0.6877 & 0.6990 & 0.7598 \\
    RIND & 0.7531 & 0.7891 & 0.7872 & 0.7665 & 0.7767 & 0.8051 \\
    UCTNet & 0.7569 & 0.7670 & 0.8201 & 0.7667 & 0.7925 & 0.8110 \\
    SFIAN & 0.7290 & 0.7513 & 0.7715 & 0.7367 & 0.7537 & 0.7896 \\
    CT-crackseg & 0.7940 & 0.7996 & \underline{0.8202 } & 0.8195 & 0.8199 & 0.8301 \\
    DTrCNet & 0.7987 & 0.8073 & 0.7972 & \underline{0.8441 } & 0.8202 & 0.8331 \\
    Crackmer & 0.7429 & 0.7640 & 0.7501 & 0.7656 & 0.7578 & 0.7966 \\
    SimCrack & \underline{0.7984 } & \underline{0.8090 } & 0.8051 & 0.8371 & \underline{0.8208 } & \underline{0.8334 } \\
    \textbf{Ours} & \textbf{0.8202 } & \textbf{0.8266 } & \textbf{0.8282 } & \textbf{0.8484 } & \textbf{0.8382 } & \textbf{0.8473 } \\
    \bottomrule
  \end{tabular}}
  \label{tab:TUT_Ours}
\end{table}

\begin{figure*}[!t]
  \centering
  \includegraphics[width=\textwidth]{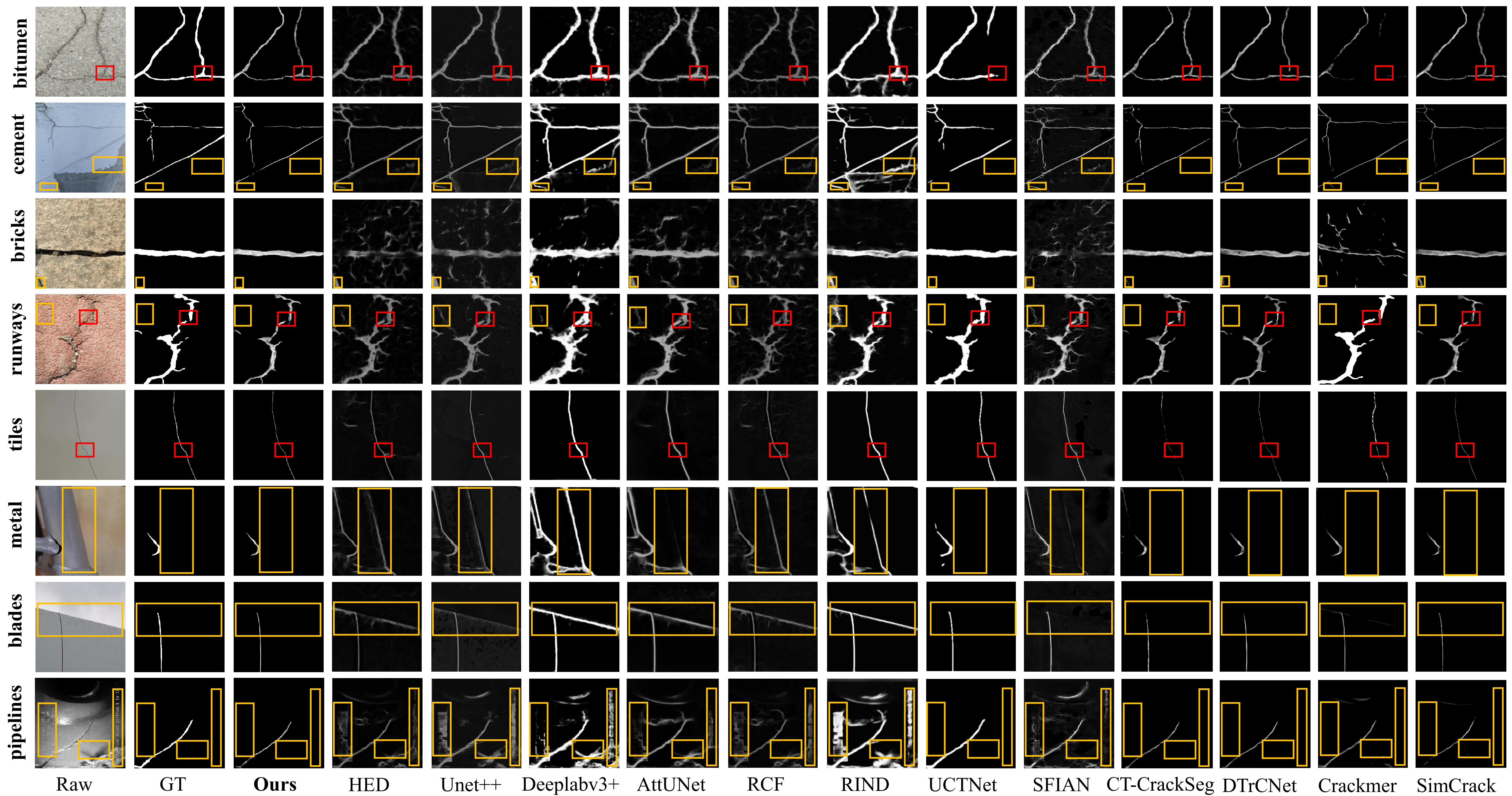}
  \caption{Visualization results of CrackSCF and other methods on TUT dataset. Key areas are highlighted with red boxes to indicate easily missed crack pixels and yellow boxes to indicate easily misdetected non-crack areas.}
  \label{fig:TUT dataset result}
  \vspace{-0.5cm}
\end{figure*}

\textit{6) Results on the TUT dataset:} 
Figure \ref{fig:TUT dataset result} presents the visual results of all methods on the TUT dataset. Easily missed crack pixel areas are highlighted with red boxes, and misdetected non-crack pixel areas are indicated with yellow boxes. As shown in Table \ref{tab:TUT_Ours}, the proposed method achieved excellent results on the TUT dataset, leading in all metrics. Specifically, it outperformed the second-best methods on ODS, OIS, P, R, F1, and mIoU by 2.73\%, 2.18\%, 0.98\%, 0.51\%, 2.12\%, and 1.67\%, respectively. While SimCrack~\cite{jaziri2024designing} achieved the second-best results on ODS, OIS, F1, and mIoU, as observed in Figure \ref{fig:TUT dataset result}, its segmentation performance on certain key details exhibits room for improvement compared to the proposed method.

Furthermore, most methods fail to suppress irrelevant background noise. In images of metal materials and turbine blades, our method effectively removes noise and accurately segments crack regions. In pipeline scenarios, our method can filter out timestamp watermarks, thanks to SCFM, which enables the model to focus on irregular, slender crack structures rather than irrelevant watermarks. Our method demonstrates excellent robustness and versatility, resulting in high-quality crack segmentation maps and adapting well to different scenarios.

\setlength\tabcolsep{6pt}
\begin{table}[!t]
    \centering
    \caption{Comparison of the complexity of CrackSCF with other methods. Best results are bolded and second best results are underlined.}
    \begin{tabular}{cccc}
    \hline
        Methods & FLOPs ↓ & Params ↓ & FPS ↑ \\ \hline
        HED & 80.35G & 14.72M & 33 \\ 
        RCF & 115.54G & 15.52M & 30 \\ 
        RIND & 695.77G & 59.39M & 11 \\ 
        UCTNet & 171.77G & 66.22M & 12 \\ 
        UNet++ & 139.61G & 9.16M & 22 \\ 
        DeeplabV3+ & 88.53G & 59.34M & 29 \\ 
        AttuNet & 541.34G & 57.16M & 24 \\ 
        SFIAN & 52.36G & 13.63M & 35 \\ 
        CT-crackseg & 39.47G & 22.88M & 28 \\ 
        DTrCNet & 123.20G & 63.45M & \textbf{47} \\ 
        SimCrack & 70.66G & 29.58M & 29 \\ 
        Crackmer & \underline{14.94G} & \underline{5.90M} & 31 \\ 

        \textbf{Ours} & \textbf{9.26G} & \textbf{4.79M} & \underline{36} \\ \hline
    \end{tabular}
    \label{tab:Complexity results}
    \vspace{-0.5cm}
\end{table}

\vspace{-0.25cm}
\subsection{Complexity Analysis}

To evaluate the efficacy of the LRDS block, its computational and parameter efficiency was analyzed using three metrics: floating point operations (FLOPs), number of parameters (Params), and frames per second (FPS). FLOPs quantify computational load, reflecting demand during inference. Params measure storage requirements, reflecting trainable and stored parameters. FPS gauges real-time processing, indicating frames processed per second under specific hardware conditions.

Table \ref{tab:Complexity results} presents our method's performance alongside other comparisons in terms of FLOPs, Params, and FPS. Our approach achieves notably low values in FLOPs (9.26G) and Params (4.79M), surpassing the second-best by 38.01\% and 18.81\%, respectively. This underscores our algorithm's efficiency, making it suitable for implementation on resource-constrained edge devices. In FPS, our model achieves the second-best performance, processing 36 frames per second, trailing only behind the speed-focused DTrCNet \cite{xiang2023crack}. This highlights our model's robust real-time capability, crucial for prompt crack image segmentation in practical applications. Overall, our model combines low parameters and computational load with competitive processing speed, ensuring effective deployment on devices like smartphones, drones, and other portable platforms.

\subsection{Ablation Studies}
To evaluate the strengths of the proposed module over other modules, ablation experiments were performed on the TUT dataset.

\begin{table}[!t]
    \centering
    \setlength\tabcolsep{6pt}
    \caption{Comparison of performance metrics using different convolution blocks. Best results are bolded and second best results are underlined.}
    \begin{tabular}{ccccccc}
    \hline
        Method & ODS & OIS & P & R & F1 & mIoU \\ \hline
        Original Conv & 0.8145 & 0.8193 & \textbf{0.8294} & 0.8398 & \underline{0.8346} & 0.8437 \\ 
        DS Conv & 0.8117 & 0.8166 & 0.8239 & 0.8364 & 0.8301 & 0.8417 \\ 
        Low-Rank Conv & \underline{0.8154} & \underline{0.8220} & 0.8220 & \underline{0.8451} & 0.8334 & \underline{0.8441} \\ 
        \textbf{LRDS Conv} & \textbf{0.8202} & \textbf{0.8266} & \underline{0.8282} & \textbf{0.8484} & \textbf{0.8382} & \textbf{0.8473} \\ \hline
    \end{tabular}
    \label{tab:conv_perf_metrics}
    \vspace{-0.5cm}
\end{table}

\begin{table}[!t]
    \centering
    \setlength\tabcolsep{6pt}
    \caption{Comparison of computational efficiency. Best results are bolded and second best results are underlined.}
    \begin{tabular}{cccc}
    \hline
        Method & FLOPs $\downarrow$ & Params $\downarrow$ & FPS $\uparrow$ \\ \hline
        Original Conv & 90.06G & 17.86M & 23 \\ 
        DS Conv & 23.91G & 8.38M & \underline{35} \\ 
        Low-Rank Conv & 21.38G & 4.63M & 30 \\ 
        \textbf{LRDS Conv} & \textbf{9.26G} & \underline{4.79M} & \textbf{36} \\ \hline
    \end{tabular}
    \label{tab:conv_efficiency_metrics}
    \vspace{-0.5cm}
\end{table}

\textit{1) Ablation study of different convolution types:}
As shown in Table \ref{tab:conv_perf_metrics}, on six metrics, our method outperforms methods that use depth-separable convolution or simple low-rank approximate convolution. As shown in Table \ref{tab:conv_efficiency_metrics}, Our method reduces FLOPs by 61.26\% and 56.69\%, respectively, and Params by 0.16M compared to the simple low-rank approximation convolution, achieving the highest FPS. Compared to original convolution operations, our method reduces FLOPs and Params by 89.70\% and 73.19\%, respectively, without degrading performance. These results indicate that LRDS effectively reduces computational load and utilizes parameters while maintaining or improving segmentation performance, enhancing processing speed and benefiting lightweight model design.

\begin{table}[!t]
    \centering
    \setlength\tabcolsep{4pt} 
    \caption{Comparison of segmentation performance  using SCFM, Single Concat branch, and Single PAF branch.}
    \begin{tabular}{ccccccc}
    \hline
        Method & ODS & OIS & P & R & F1 & mIoU \\ \hline
        Single Concat branch & 0.8019 & 0.8092 & 0.8228 & 0.8209 & 0.8219 & 0.8359 \\ 
        Single PAF branch & \underline{0.8137} & \underline{0.8202} & \underline{0.8282} & \underline{0.8297} & \underline{0.8293} & \underline{0.8427} \\ 
        \textbf{SCFM} & \textbf{0.8202} & \textbf{0.8266} & \textbf{0.8289} & \textbf{0.8484} & \textbf{0.8382} & \textbf{0.8473} \\ \hline
    \end{tabular}
    \label{tab:scfm_ablation_metrics}
    \vspace{-0.5cm}
\end{table}

\begin{table}[!t]
    \centering
    \setlength\tabcolsep{8pt} 
    \caption{Comparison of computational efficiency. Best results are bolded and second best results are underlined.}
    \begin{tabular}{cccc}
    \hline
        Method & FLOPs $\downarrow$ & Params $\downarrow$ & FPS $\uparrow$ \\ \hline
        Single Concat branch & \textbf{7.21G} & \textbf{4.71M} & \textbf{42} \\ 
        Single PAF branch & \underline{8.24G} & \underline{4.72M} & \underline{39} \\ 
        \textbf{SCFM} & 9.26G & 4.79M & 36 \\ \hline
    \end{tabular}
    \label{tab:scfm_ablation_efficiency}
        \vspace{-0.5cm}
\end{table}

\textit{2) Ablation study in SCFM using different branch:}
To demonstrate the effectiveness of SCFM in fusing local detail features and long-range information, ablation experiments were performed on the TUT dataset, As shown in Table \ref{tab:scfm_ablation_metrics} and \ref{tab:scfm_ablation_efficiency}. Our method, using Concat and PAF dual-branch feature fusion, achieves the best results across all six segmentation metrics. Using only the PAF branch results in a 1.06\% drop in the F1 score. While SCFM slightly increases parameter count, computational load, and inference speed, it significantly improves segmentation performance. These results indicate that our module effectively balances performance and efficiency.

\begin{table}[!tp]
    \centering
    \setlength\tabcolsep{4pt}
    \caption{Comparison of segmentation performance of SCFM with other segmentation heads. Best results are bolded and second best results are underlined.}
    \begin{tabular}{ccccccc}
    \hline
        SegHead & ODS & OIS & P & R & F1 & mIoU \\ \hline
        PID & 0.7120 & 0.7198 & \underline{0.7921} & 0.6742 & 0.7284 & 0.7777 \\ 
        SegFormer & 0.7446 & 0.7634 & 0.7533 & 0.7786 & 0.7658 & 0.7983 \\ 
        UNet & \underline{0.7836} & \underline{0.7951} & 0.7829 & 0.8248 & \underline{0.8033} & \underline{0.8222} \\ 
        Hamburger & 0.7447 & 0.7609 & 0.7075 & \underline{0.8328} & 0.7670 & 0.7969 \\ 
        \textbf{SCFM} & \textbf{0.8202} & \textbf{0.8266} & \textbf{0.8289} & \textbf{0.8484} & \textbf{0.8382} & \textbf{0.8473} \\ \hline
    \end{tabular}
    \label{tab:seghead_perf}
    \vspace{-0.5cm}
\end{table}

\begin{table}[!tp]
    \centering
    \setlength\tabcolsep{8pt}
    \caption{Comparison of computational efficiency of SCFM with other segmentation heads. Best results are bolded and second best results are underlined.}
    \begin{tabular}{cccc}
    \hline
        SegHead & FLOPs $\downarrow$ & Params $\downarrow$ & FPS $\uparrow$ \\ \hline
        PID & 36.69G & \underline{5.44M} & 21 \\ 
        SegFormer & \underline{29.72G} & 6.20M & \underline{23} \\ 
        UNet & 40.59G & 12.34M & 20 \\ 
        Hamburger & 168.23G & 7.45M & 14 \\ 
        \textbf{SCFM} & \textbf{9.26G} & \textbf{4.79M} & \textbf{36} \\ \hline
    \end{tabular}
    \label{tab:seghead_efficiency}
    \vspace{-0.5cm}
\end{table}

\textit{3) Ablation study of different segmentation heads:} To demonstrate the advantages of SCFM over other high-performance segmentation heads, ablation experiments were conducted on different segmentation heads, including PID \cite{xu2023pidnet}, SegFormer \cite{xie2021segformer}, UNet \cite{ronneberger2015u}, and Hamburger head \cite{geng2021attention}, using the TUT dataset. As shown in the Table \ref{tab:seghead_perf}, SCFM achieves the best results in segmentation performance metrics. Particularly in terms of F1 and mIoU, SCFM outperforms the UNet head, which achieved the second-best results, by 4.35\% and 3.05\%, respectively. This indicates that SCFM effectively enhances the model's segmentation performance by perceiving and integrating local details and long-range dependencies of crack pixels. Furthermore, as shown in \ref{tab:seghead_efficiency}, SCFM requires low computational resources, with Params and FLOPs being 11.95\% and 68.84\% lower than those of the second-best results, respectively, and achieves the highest FPS. This demonstrates that our SCFM maintains excellent performance while requiring low computational resources.

\begin{table}[!t]
    \centering
    \setlength\tabcolsep{4pt}
    \caption{Comparison of segmentation performance using our lightweight deformable attention in LDE with other variants. Best results are bolded and second best results are underlined.}
    \begin{tabular}{ccccccc}
    \hline
        Attention & ODS & OIS & P & R & F1 & mIoU \\ \hline
        Common MHA & 0.7919 & 0.8009 & 0.7999 & 0.8240 & 0.8118 & 0.8279 \\ 
        MHA + LR & 0.7924 & 0.8003 & 0.7988 & 0.8284 & 0.8134 & 0.8286 \\ 
        Common MSDA & \underline{0.8058} & \underline{0.8151} & \underline{0.8052} & \underline{0.8467} & \underline{0.8257} & \underline{0.8373} \\ 
        \textbf{MSDA} & \textbf{0.8202} & \textbf{0.8266} & \textbf{0.8289} & \textbf{0.8484} & \textbf{0.8382} & \textbf{0.8473} \\ \hline
    \end{tabular}
    \label{tab:attn_perf}
    \vspace{-0.5cm}
\end{table}

\begin{table}[!t]
    \centering
    \setlength\tabcolsep{8pt}
    \caption{Comparison of computational efficiency with different attention variants. Best results are bolded and second best results are underlined.}
    \begin{tabular}{cccc}
    \hline
        Attention & FLOPs $\downarrow$ & Params $\downarrow$ & FPS $\uparrow$ \\ \hline
        Common MHA & 15.98G & 5.80M & 16 \\ 
        MHA + LR & \textbf{8.87G} & \textbf{4.77M} & 14 \\ 
        Common MSDA & 16.37G & 5.82M & \underline{28} \\ 
        \textbf{MSDA + LR} & \underline{9.26G} & \underline{4.79M} & \textbf{36} \\ \hline
    \end{tabular}
    \label{tab:attn_efficiency}
    \vspace{-0.5cm}
\end{table}

\textit{4) Ablation study of different attention variants in LDE:} To validate that the LR Linear can effectively reduce the computational resources required by the deformable attention (MSDA) in LDE while achieving excellent segmentation performance, lightweight MSDA was compared with common multi-head attention (MHA), MHA with embedded LR Linear, and common MSDA. As shown in Table \ref{tab:attn_perf} and \ref{tab:attn_efficiency}, the proposed method achieves the best results in segmentation performance metrics. Specifically, compared to the common MSDA, the proposed method improves by 1.51\% and 1.19\% in F1 and mIoU, respectively. Compared to MHA, MSDA outperforms it in segmentation performance, demonstrating the effectiveness of utilizing MSDA. Furthermore, whether for MHA or MSDA, when embedded with LR Linear, the required Params and FLOPs significantly decrease. For MSDA, the Params and FLOPs are reduced by 43.43\% and 17.70\%, respectively. This demonstrates that the LR Linear can effectively reduce the computational requirements needed by the attention mechanism.

\setlength\tabcolsep{6pt}
\begin{table}[!t]
    \centering
    \caption{Comparison of results of CrackSCF networks trained with different loss ratios. Best results are bolded and second best results are underlined}
    \begin{tabular}{ccccccc}
    \hline
        Dice : BCE & ODS & OIS & P & R & F1 & mIoU \\ \hline
        1 : 5 & 0.8129  & 0.8203  & 0.8217  & \underline{0.8458}  & \underline{0.8336}  & 0.8424  \\ 
        1 : 4 & 0.8049  & 0.8145  & 0.8116  & 0.8365  & 0.8239  & 0.8368  \\ 
        1 : 3 & \textbf{0.8202}  & \textbf{0.8266}  & 0.8282  & \textbf{0.8484}  & \textbf{0.8382}  & \textbf{0.8473}  \\ 
        1 : 2 & \underline{0.8146}  & \underline{0.8211}  & 0.8227  & 0.8435  & 0.8330  & \underline{0.8438}  \\ 
        1 : 1 & 0.8136  & 0.8204  & 0.8231  & 0.8436  & 0.8332  & 0.8431  \\ 
        Only BCE & 0.8003  & 0.8056  & \textbf{0.8349}  & 0.8004  & 0.8173  & 0.8332  \\ 
        Only Dice & 0.8107  & 0.8165  & \underline{0.8302}  & 0.8265  & 0.8283  & 0.8409  \\ \hline
    \end{tabular}
    \label{tab:Abl Loss results}
\end{table}

\textit{5) Analysis of different hyperparameter settings in the loss function:} 
Table \ref{tab:Abl Loss results} shows the results of ablation experiments on different weight proportions of the two sub-losses in the loss function. As shown, a Dice loss to BCE loss ratio of 1:3 yields the highest values in five metrics: ODS, OIS, R, F1, and mIoU. Although the P metric is 0.80\% lower than using only BCE loss, the F1 metric improves by 2.58\%, indicating a better balance between P and R. Therefore, this ratio is adopted during training. In summary, a Dice loss to BCE loss ratio of 1:3 not only enhances pixel-level performance but also improves the global segmentation of crack areas, achieving superior segmentation performance.

\subsection{Limitations and Discussions}
Through these experiments, it is evident that SCFM effectively perceives and fuses local details and long-range pixel dependencies. The proposed network excels in various metrics on public datasets with simple scenes and performs exceptionally on the complex TUT dataset. The lightweight LRDS block minimizes parameters and computational load, facilitating deployment on resource-constrained devices like drones and smartphones. Additionally, LDE is efficient in capturing irregular long-distance dependencies between cracked pixels. However, the proposed network presents several limitations, as follows:

\begin{figure}[!t]
  \centering
  \includegraphics[width=0.7\textwidth]{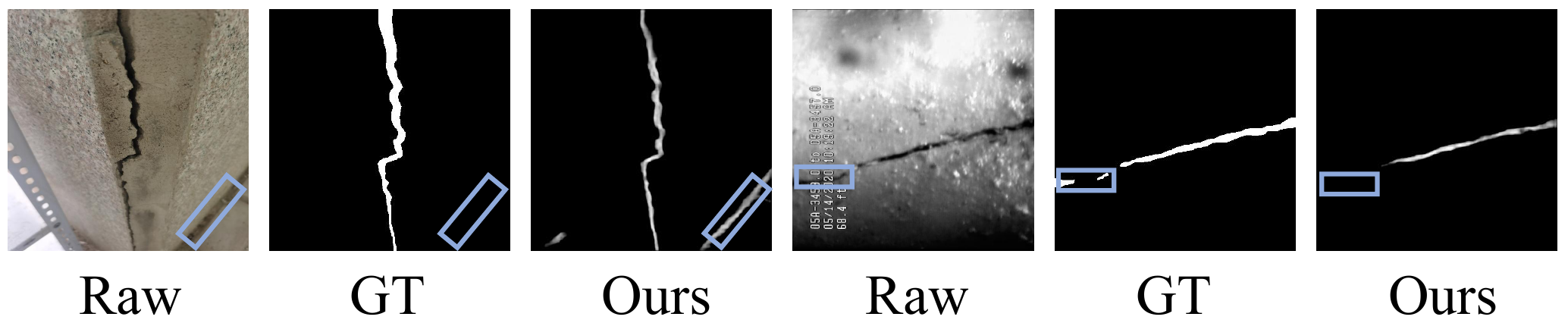}
  \caption{Examples of CrackSCF misdetections and misses. The critical regions are marked with blue boxes.}
  \label{fig:Limitation}
\end{figure}

\begin{enumerate}

\item Although our method generally achieves good crack segmentation with effective background noise suppression, as shown in the Figure \ref{fig:Limitation}, our model occasionally misdetects noise regions that resemble cracks in some images. It also fails to detect watermark occlusions with colors similar to the background. Therefore, it is essential to enhance modules such as MFE and LDE to improve the model's capability to perceive and extract features from critical noise and occlusion areas, thereby further increasing its robustness and stability in complex scenarios.

\item Despite our method achieves the minimum parameter count and computational load, there is still room for improvement in inference speed. For example, our network achieves a processing speed of 36 frames per second, but the DTrCNet \cite{xiang2023crack}, which focuses more on inference speed, attains a faster speed of 47 frames per second. Therefore, it is necessary to further simplify convolution operations or the network's complexity to achieve faster inference speeds while maintaining a good balance between segmentation performance and model lightweighting.

\end{enumerate}

\section{Conclusion}
\label{sec:conclusion}

This paper proposes a staircase cascaded fusion crack segmentation network (CrackSCF) for pixel-level crack detection. Within CrackSCF, the lightweight convolutional block LRDS is introduced to construct the MFE. This block dramatically reduces the computational load and parameter quantity required by the model while efficiently capturing the local details of crack images. Additionally, the LDE is proposed to more efficiently perceive irregular long-range dependencies among pixels in the crack feature maps. To comprehensively fuse local patterns with these long-range dependencies, the SCFM is employed to generate higher-quality segmentation maps. Furthermore, to comprehensively evaluate the model's robustness and stability across various complex scenarios, the TUT dataset, which includes eight types of crack scenes, was collected. Experimental results show that the proposed method performs best on five public datasets and the TUT dataset, particularly excelling in background noise suppression and fine crack segmentation. The computational load and parameter count are reduced to 9.26G and 4.79M, respectively.

Although the proposed method achieves a low parameter count and computational load, there is still room for improvement in inference speed. In future work, network lightweighting will be further pursued to reduce computational resource requirements and achieve higher inference speeds, with the aim of deploying the model on mobile devices for enhanced real-world applications. Additionally, future efforts will focus on designing an improved SCFM to better fuse and enhance local information and global context for superior segmentation results.









\bibliographystyle{elsarticle-num-names_1} 
\bibliography{References_1}


\end{document}